\pdfoutput=1
\documentclass[journal]{IEEEtran}
\usepackage{graphicx}
\usepackage{amsmath,amssymb}
\usepackage{lineno}
\usepackage{color}
\usepackage{subfigure}
\usepackage{txfonts}
\usepackage{multirow}
\usepackage{algorithm}
\begin{document}
\title{Describing Human Aesthetic Perception by Deeply-learned Attributes from Flickr}
\author{Luming Zhang \IEEEcompsocitemizethanks{\IEEEcompsocthanksitem L. Zhang is with the School of Computing, National
University Singapore.}} \maketitle
\markboth{A Submission to IEEE Transactions on Multimedia}%
{Shell \MakeLowercase{\textit{et al.}}: Bare Demo of IEEEtran.cls
for Journals} \maketitle

\begin{abstract}
Many aesthetic models in computer vision suffer from two
shortcomings: 1) the low descriptiveness and interpretability of
those hand-crafted aesthetic criteria (\textit{i.e.},
non-indicative of region-level aesthetics), and 2) the difficulty
of engineering aesthetic features adaptively and automatically
toward different image sets. To remedy these problems, we develop
a deep architecture to learn aesthetically-relevant visual
attributes from Flickr\footnote{https://www.flickr.com/}, which
are localized by multiple textual attributes in a
weakly-supervised setting. More specifically, using a bag-of-words
(BoW) representation of the frequent Flickr image tags, a
sparsity-constrained subspace algorithm discovers a compact set of
textual attributes (\textit{e.g.}, landscape and sunset) for each
image. Then, a weakly-supervised learning algorithm projects the
textual attributes at image-level to the highly-responsive image
patches at pixel-level. These patches indicate where humans look
at appealing regions with respect to each textual attribute, which
are employed to learn the visual attributes. Psychological and
anatomical studies have shown that humans perceive visual concepts
hierarchically. Hence, we normalize these patches and feed them
into a five-layer convolutional neural network (CNN) to mimick the
hierarchy of human perceiving the visual attributes. We apply the
learned deep features on image retargeting, aesthetics ranking,
and retrieval. Both subjective and objective experimental results
thoroughly demonstrate the competitiveness of our approach.
\end{abstract}


\section{Introduction}
Perceptually aesthetic modeling refers to the process of
discovering low-level and high-level visual patterns that can
arouse human aesthetic perception. It is a useful technique to
enhance many applications, such as image retargeting, photo album
management and scene rendering. Take retargeting for example, the
aesthetically pleasing image regions are squeezed slightly and
vice versa. Moreover, effectively describing region-level
aesthetics can guide the level of details in non-photorealistic
image rendering. As a popular media sharing website with millions
of photos uploaded and commented daily, Flickr can be considered
as an ideal platform to study and simulate how humans perceive
photos with aesthetic features at both low-level and high-level.
However, perceptually aesthetic modeling based on Flickr will
encounter the following challenges:\begin{figure*}[htp!]\centering
\includegraphics[scale=0.55]{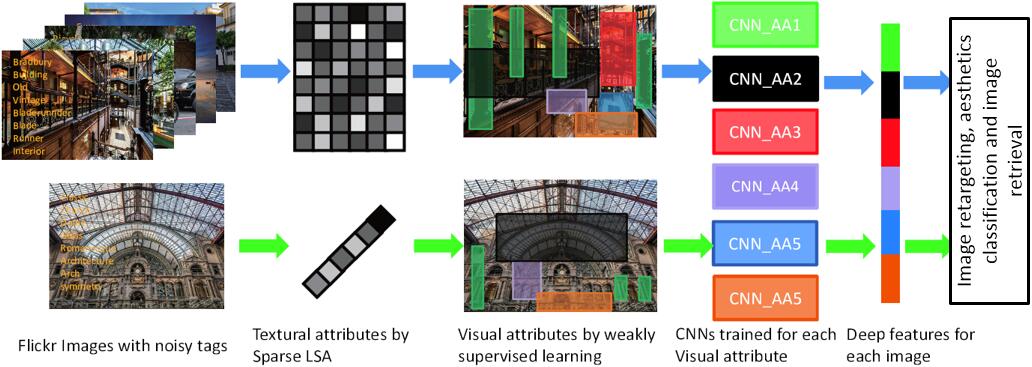}
\caption{The pipeline of the proposed CNN-based aesthetic modeling
framework (The blue and green arrows denote the training and test
phases respectively. The color-tagged regions indicate visual
attributes which are aesthetically
pleasing.).}\label{fig1}\end{figure*}\begin{itemize} \item Flickr
contains a large amount of photos with multiple latent aesthetic
attributes (\textit{e.g.}, the ``field guide'' and ``movement'').
Conventional aesthetic models, however, are typically built upon
generic features which quantify the compliance to a pre-defined
criterion (\textit{e.g.}, the ``rule of thirds''). In practice, we
need tailored and dataset-dependent features to capture these
latent aesthetic attributes, but engineering these features may
require the domain knowledge of professional photographers. \item
The interpretability of an aesthetic model means the capability of
indicating which regions are responsive to each aesthetic
attribute. A well interpretable aesthetic model has both
scientific impact and application value. As far as we know, the
existing aesthetic modeling pipeline is more or less a blackbox.
They have limited power to provide region-level response to each
aesthetic attribute, particularly the abstract and global ones
such as ``harmonic'', ``dream'', and ``colorful''. \item Previous
methods associate each photo with a set of aesthetic attributes in
well-controlled settings. For example, the training images'
beautifulness is quantified to reduce the disturbance of noisy
tags~\cite{ava}. In reality, image tags from Flickr are labeled
uncontrollably by a variety of users with diverse backgrounds,
education,~\textit{etc}. This will inevitably produces noisy image
tags, which necessitate an aesthetic model to robustly handle
noises from image tags.\end{itemize} To solve these problems, we
propose a CNN-based framework which models aesthetic perception
automatically and interpretably, by calculating a compact set of
textual and visual attributes from tagged Flickr images. The
pipeline of our proposed framework is presented in
Fig.~\ref{fig1}. For each tagged image from a large-scale Flickr
corpus, we employ a bag-of-words (BoW) representation of those
frequently-occurring and noisy image tags. Toward an efficient
system, a sparsity-constrained subspace algorithm projects each
BoW histogram onto a compact set of textual attributes. They
reflect the highly representative aesthetic attributes of an
image. To locate the appealing visual attributes, a weakly
supervised algorithm maps the textual attributes at image-level
onto the salient patches (indicated by different colors in the
middle of Fig.~\ref{fig1}) in an image. Both psychological and
anatomical studies have shown that human vision system is
multi-layered and forms higher-level abstracts from input raw
pixels incrementally. That means the intrinsic hierarchical
structure of CNN has the potential to model human visual
perception. Based on this, for each textual attribute, the
corresponding extracted patches are employed to learn a five-layer
CNN to simulate the hierarchial perception of human beings. The
learned deep feature is applied on a range of multimedia
tasks: image retargeting, aesthetics classification and image retrieval.\\
\indent The contributions of this paper can be summarized as: 1)
The first deep architecture that learns aesthetically-relevant
visual attributes extracted from massive-scale Flickr images; 2) a
weakly supervised algorithm associating each textual attribute
with the corresponding visual attribute; and 3) adopting the deep
features on several multimedia applications, combined with
extensive experimental validation.\\
\indent The rest of this paper is organized as follows: Sec II
briefly reviews the related work. Sec III introduces the proposed
deep image aesthetics modeling framework, including the
sparsity-constrained textual attributes mining, weakly supervised
visual attributes detection, our developed CNN and it
applications. Experimental results in Sec IV thoroughly
demonstrate the effectiveness of our method. Sec V concludes.

\section{Related Work}
Our work is related to two research topics in multimedia
domain~\cite{tmm1,tmm2}: computational image aesthetics analysis
and deep learning-based aesthetic attribute modeling.
\subsection{Computational Image Aesthetic Analysis}
\subsubsection{Global features-based models} Datta~\textit{et
al.}~\cite{datta} proposed 58 low-level visual
features,~\textit{e.g.}, the shape convexity, to capture photo
aesthetics. Dhar~\textit{et al.}~\cite{dhar} proposed a set of
high-level attribute-based predictors to evaluate photo
aesthetics. In~\cite{luo2}, Luo \textit{et al.} employed a
GMM-based hue distribution and a prominent line-based texture
distribution to assess the global composition of an image. To
represent image local composition, regional features describing
human faces, region clarity and region complexity were developed.
In~\cite{marchesotti}, Marchesotti~\textit{et al.} tackled the
problem of visual aesthetic modeling by discovering mid-level
features. The designed algorithm can automatically learn semantic
aesthetics-related attributes by combining image, scoring, and
textual data from the AVA data set. The authors shown that the
learned attributes can facilitate a variety of media
applications,~\textit{e.g.}, aesthetic quality prediction, image
tagging and retrieval. Experimental results shown that the above
two generic descriptors outperform a variety
of hand-crafted and dataset-dependent aesthetic descriptors.
\subsubsection{Local feature-based models} Cheng~\textit{et
al.}~\cite{cheng} proposed the omni-range context,~\textit{i.e.},
the spatial distribution of arbitrary pairwise image patches, to
describe image composition. Nishiyama~\cite{nishiyama}~\textit{et
al.} first detected multiple subject regions in a photo, and
afterward an SVM classifier is trained for each subject region.
Finally, the aesthetics of an image is quantified by combining the
SVM scores of a photo's internal subject regions.
In~\cite{nishiyama2}, Nishiyama~\textit{et al.} proposed a color
harmony-based aesthetic model, which models image color
distribution by leveraging its patches. The patch-level color
distribution is integrated into a BoW histogram, which is
classified by an SVM to determine whether a photo is highly or low
aesthetic. Bhattacharya~\textit{et al.}~\cite{bhattacharya}
developed a spatial recomposition system which allows users to
select a foreground object interactively. The system presents
recommendations to indicate an optimal location of the foreground
object, which is detected by integrating multiple visual features.
\subsection{Deep learning-based Aesthetic Attribute Modeling}
As far as we know, there are only two deep learning models for
visual aesthetic analysis. In~\cite{da9}, Lu~\textit{et al.}
proposed a double-layer CNN architecture to automatically discover
effective features that capture image aesthetics from two
heterogeneous input sources, i.e., aesthetic features from both
the global and local views. The double CNN layers are jointly
trained from two inputs. The first layer takes global image
representation as the input, while the second layer takes local
image representations as the input. This allows us to leverage
both compositional and local visual information. Based on the
evaluation from the AVA data set, Lu ~\textit{et al.}'s algorithms
significantly outperforms the results reported earlier. This model
differs from ours fundamentally in two aspects: 1) Both the global
and local views are heuristically defined, there is no guarantee
that they can well locate the aesthetically pleasing regions
across different datasets. Comparatively, our approach uses a
weakly supervised algorithm to discover visually appealing regions
indicated by tags. Thus, it can be conveniently adapted onto
different datasets; 2) Lu~\textit{et al.}'s model simply captures
the global and local aesthetic features of a photo. But there is
no evidence that abstract aesthetic cues such as ``vivid'' and
``harmonic'' can be well described. Noticeably, in our model, a
set of CNNs are trained. Each encodes visual attribute
corresponding to a textual attribute, which can capture either a
concrete or abstract aesthetic cue. In~\cite{da10},
Champbell~\textit{et al.} trained two Restricted Boltzmann
Machines (RBMs) on highly and low aesthetic images respectively.
The authors observed that 10\% of the filters learned from the
highly aesthetic images capture the aesthetics-relevant visual
cues. But this model is only available on simple abstract
paintings with low resolution. It is computationally intractable
to describe high resolution Flickr images with various semantics.

\section{The Proposed Method}
\subsection{Sparse Textual Attributes Discovery} Given a Flickr
image, we use an $M$-dimensional augmented frequency vector
$\overrightarrow{\alpha}$ to represent the distribution of its
tags\footnote{In our implementation, we set $M=100$.}. In
particular, to avoid the randomly-occurring noisy image tags, we
select the $M$ most frequent tags from the training image set.
Then, we treat the tag set of each Flickr image as a document
$\mathcal{D}$, based on which the $i$-th element of vector
$\overrightarrow{\alpha}$ can be calculated as:
\begin{equation} \overrightarrow{\alpha}(i)=0.5+\frac{0.5*f(i,\mathcal{D})}
{\max(f(j,\mathcal{D}):j\in \mathcal{D})}, \label{eq1}
\end{equation}
where $f(i,\mathcal{D})$ counts the times of the $i$-th tag from
the $M$ most frequent ones occurring in document $d$, and the
denominator functions as a normalization factor. In our
implementation, we
set $M=100$ based on cross validation.\\
\indent Given $N$ Flickr images, they can be represented as $N$
augmented frequency vectors. Thereafter, we column-wise stack them
into a matrix
$\mathbf{X}=[\overrightarrow{\alpha}_1;\overrightarrow{\alpha}_2;\cdots;\overrightarrow{\alpha}_N]\in
\mathbb{R}^{N\times M}$, where each row $\vec{X}_j\in
\mathbb{R}^N$ denotes the $j$-th feature vector cross all the
documents. To obtain the textual attributes of each Flickr image,
we adopt a subspace algorithm, which converts the original
$M$-dimensional vector corresponding to each Flickr
image into a $D$-dimensional textual attribute vector ($D<\min(M,N)$).
In our implementation, all the 75246 training images are adopted to create document matrix $\mathbf{X}$\\
\indent Following the latent semantic analysis (LSA)~\cite{lsa},
we assume that the $D$ textual attributes
$\{\vec{u}_1,\vec{u}_2,\cdots,\vec{u}_D\}$ are uncorrelated, where
each attribute $\vec{u}_d\in \mathbb{R}^N$ has the unit
length,~\textit{i.e.}, $||\vec{u}_d||_2=1$. Denote
$\mathbf{U}=[\vec{u}_1,\vec{u}_2,\cdots,\vec{u}_D]\in
\mathbb{R}^{N\times D}$, we have
$\mathbf{U}^T\mathbf{U}=\mathbf{I}$ where $\mathbf{I}$ is the
identity matrix. It is reasonable to assume that each feature
vector $\vec{X}_j$ can be linearly reconstructed by the textual
attributes:
\begin{equation}
\mathbf{X}_j=\sum\nolimits_{d=1}^D
a_{dj}\mathbf{u}_d+\epsilon_j,\label{eq2}
\end{equation}
In the matrix form, the above equation can be reorganized into
$\mathbf{X}=\mathbf{UA}+\mathbf{\epsilon}$, where
$\mathbf{A}=[a_{dj}]\in \mathbb{R}^{D\times M}$ denotes the
projection matrix from the tag space to the textual attribute
space. We can obtain the projection matrix $\mathbf{A}$ by solving
the following optimization. It minimizes the rank-$D$
approximation error subject to the orthogonality constraint of
$\mathbf{U}$:\begin{equation}
\min\nolimits_{\mathbf{U},\mathbf{A}}
||\mathbf{X}-\mathbf{UA}||_F^2,~~~~s.t.~~~~\mathbf{U}^T\mathbf{U}=\mathbf{I},\label{eq3}
\end{equation}
where $||\cdot||_F$ denotes the matrix Frobenius norm. The
constraint $\mathbf{UU}^T=\mathbf{I}$ reflects the uncorrelated
property among textual attributes.\\
\indent \textbf{Sparsity of textual attributes:} The projection
matrix $\mathbf{U}$ learned from (\ref{eq3}) can reconstruct
matrix $\textbf{X}$ by a linear combination of textual attributes.
Typically, the number of textual attributes $D$ is not
small\footnote{As experimented on our own complied data set, each
Flickr image is associated with $2\sim 5$ textural attributes and
the optimal total number of attributes $D$ is between 10 and 15.}.
It means that aesthetically modeling an Flickr image by analyzing
all its correlated textual attributes might be intractable. Toward
an efficient aesthetic model, we encode a sparsity constraint into
(\ref{eq3}) in order to achieve a sparse projection matrix
$\mathbf{A}$. An entry-wise $l_1$-norm of $\mathbf{A}$ is added as
a regularization term to the loss function. Based on this, we
formulate the sparse textual attributes discovery as:
\begin{equation}
\min\nolimits_{\mathbf{U},\mathbf{A}}
||\mathbf{X}-\mathbf{UA}||_F^2+\lambda||\mathbf{A}||_1,~~s.t.,
~~\mathbf{U}^T\mathbf{U}=\mathbf{I},\label{eq4}
\end{equation}
where $||\mathbf{A}||_1=\sum\nolimits_{d=1}^D
\sum\nolimits_{j=1}^M|a_{dj}|$ is the entry-wise $l_1$-norm of
$\mathbf{A}$; and $\lambda$ is the positive regularization
parameter controlling the density of $\mathbf{A}$,~\textit{i.e.},
the number of nonzero entries. In general, a larger $\lambda$
leads to a sparser $\mathbf{A}$. On the other hand, a too sparse
$\mathbf{A}$ will lose some relationships between Flickr tags and
textual attributes and will in turn harm the reconstruction
quality. In practice, it is important to select an appropriate
$\lambda$ to obtain a more sparse $\mathbf{A}$ while still
achieving a good reconstruction performance. It is noticeable that
(\ref{eq4}) is solved by alternatively optimizing matrix
$\mathbf{U}$ and $\mathbf{A}$, as detailed in~\cite{slsa}.
\subsection{Weakly Supervised Visual Attributes Learning} This
section introduces a graphlet-based weakly supervised learning
framework to detect visual attributes corresponding to each
textual attribute. We first construct a superpixel pyramid which
captures objects with different shapes seamlessly. To quantify
region-level response to each textual attribute, a manifold
embedding algorithm transfers textual attributes into graphlets.
Finally, the patch containing graphlet most responsive to each
textural attribute form the visual attribute.\\
\indent \textbf{Graphlet construction:} There are usually millions
of raw pixels in an image, treating each of them independently
brings intractable computation. It is generally accepted that
pixels within an images are highly correlated with its spatial
neighbors. Therefore, we sample a collection of superpixels and
use them to construct different objects. As objects with different
scales may evoke human aesthetic perception, we construct
superpixels with three sizes by overly, moderately and deficiently
segmenting each image. The segmentation is based on the well-known
simple linear iterative clustering (SLIC). It is built upon the
k-means clustering and has a time complexity of $\mathcal{O}(N)$,
where $N$ is the number of pixels in an image. Compared with the
conventional methods, experiments shown that SLIC is
computationally more efficient, requires less memory and generates
superpixels more adherent to object boundaries. By segmenting each
image three times with different SLIC parameters, a superpixel
pyramid is constructed to capture objects with different
sizes.\\\begin{figure}[htp!]\centering
\includegraphics[scale=0.32]{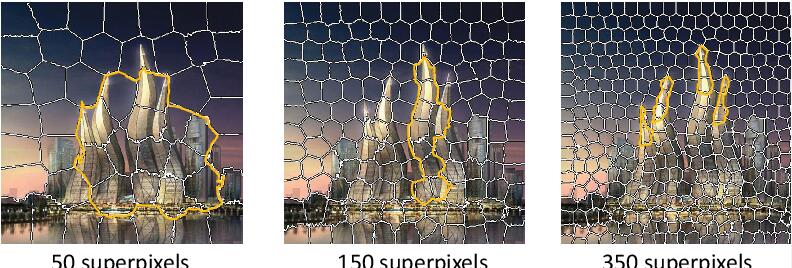}
\caption{A superpixel pyramid describing the Dubai towers from
coarse to fine. The first pyramid layer describes the rough
outline of the four towers, the surrounding lake and the sky,
reflecting textual attributes such as ``harmonic'' and
``symmetric''. The second layer captures the shape of each
individual tower, which corresponds to textual attributes such as
``flame'' and ``dynamic''. The third layer encodes the details of
the four towers,~\textit{e.g.}, the shape tower top. It is highly
responsive to textual attributes such as ``shape'' and
``edge''.}\label{fig2}\end{figure} \indent As shown in
Fig.~\ref{fig2}, different objects can be constructed by a set of
spatially neighboring superpixels. More specifically, a graphlet
is a moderately sized graph defined as:
\begin{equation}
G=(V,E),\label{eq11}
\end{equation}
where $V$ is a set of vertices, each representing a superpixel;
$E$ is a set of edges, each connecting a pair of spatially
neighboring superpixels. We call a graphlet $t$-sized if it is
constituted by $t$ superpixels.\\
\indent Given a $t$-sized graphlet, we represent it by a $t\times
(t+128+9)$ matrix as:
\begin{equation}
\mathbf{M}=[\mathbf{M}_C,\mathbf{M}_T,\mathbf{M}_S],\label{eq12}
\end{equation}
where $\mathbf{M}_C$ is a $t\times 9$ matrix and each row is the
9-dimensional color moment~\cite{cm} from a superpixel;
$\mathbf{M}_T$ is a $t\times 128$ matrix where each row is a
128-dimensional HOG~\cite{hog} from a superpixel; and
$\mathbf{M}_S$ is a $t\times t$ adjacency matrix representing the
topology of a graphlet. The graphlet extraction is based on random
walk on the superpixel mosaic. More specifically, we first index
the superpixels and then select a starting one. Afterward,
spatially neighboring superpixels are visited one-by-one until the
maximum graphlet size is reached. Due to the number of graphlets
is exponentially increasing the the maximum graphlet size, we set
it to 7 in our implementation.\\
\indent \textbf{Weakly supervised semantic encoding:} The textual
attributes indicate the aesthetically pleasing regions in a Flickr
image. To locate them, we propose a weakly supervised learning
algorithm which transfers textual attributes into different
graphlets in an image. The objective function is:
\begin{equation} \arg\min_{\mathbf{Y}}\sum_{ij}
||y_i-y_j||^2[l_s(i,j)-l_d(i,j)],~s.t.,~\mathbf{YY}^T=\mathbf{I}_d,
\label{eq13}\end{equation} where
$\mathbf{Y}=[y_1,y_2,\cdots,y_n]$, each denoting a $d$-dimensional
post-embedding graphlet from the training images. $\mathbf{Y}$ in
Eq.(7) is initialized by the the vector obtaining by row-wise
concatenating matix $\mathbf{M}$ in Eq.(6). The number of
graphlets used in $\mathbf{Y}$ depends on different image sets. In
our experiment, the number is 75246. $l_s$ and $l_d$ are functions
measuring the semantic similarity and difference between
graphlets, which are quantified according to textual attributes.
Denoting $\overrightarrow{n}=[n^1,n^2,\cdots,n^C]^T$ where $n^i$
is the number of images with the $i$-th textual attribute, and
$c(\cdot)$ contains the textual attributes of the Flickr image
from which a graphlet is extracted, then we can obtain:
\begin{equation} l_s(i,j)=\frac{[c(G_i)\cap
c(G_j)]\overrightarrow{n}}{\sum\nolimits_c n^c},\label{eq14}
\end{equation}
\begin{equation}
l_d(i,j)=\frac{[c(G_i)\oplus
c(G_j)]\overrightarrow{n}}{\sum\nolimits_c n^c},\label{eq15}
\end{equation}
where the numerator of $l_s$ denotes the number of images sharing
the common textual attributes with the images where the $i$-th and
$j$-th graphlets are extracted; the numerator of $l_d$ is the
number of images having different textual attributes with the
images where the $i$-th and $j$-th graphlets are extracted. The
denominator represents the total number of images with all textual
attributes, which is used as a normalization factor. An example of
calculating $l_s$ and $l_d$ is presented in Fig.~\ref{fig3}, where
$l_s(i,j)=\frac{5+3+7}{5+1+\cdots+1}=0.333$ and
$l_d(i,j)=\frac{1+3+1+4+3}{5+1+\cdots+1}=0.200$.\\
\indent Objective function (\ref{eq13}) can be reorganized into:
\begin{eqnarray}
&&\arg\min\nolimits_{\mathbf{Y}}\sum\nolimits_{ij}||y_i-y_j||^2[l_s(i,j)-l_d(i,j)]\nonumber\\
&&=\arg\max\nolimits_{\mathbf{Y}}\text{tr}(\mathbf{YUY}^T),~~s.t.~~\mathbf{YY}^T=\mathbf{I}\label{eq16}
\end{eqnarray}
where $\mathbf{U}=[-\overrightarrow{e}_{N-1}^T,\mathbf{I}_{N-1}]^T\mathbf{W}_1
[-\overrightarrow{e}_{N-1}^T,\mathbf{I}_{N-1}]+\cdots+[\mathbf{I}_{N-1},
-\overrightarrow{e}_{N-1}^T]^T\mathbf{W}_N[\mathbf{I}_{N-1},-\overrightarrow{e}_{N-1}^T]$,
and $\mathbf{W}_1$ is an $N\times N$ diagonal matrix whose $h$-th diagonal element is
$l_s(h,i)-l_d(h,i)$. Note that (\ref{eq16}) is a quadratic problem with quadratic constraints, which can be solved
by eigenvalue decomposition with a time complexity of $\mathcal{O}(N^3)$. To speedup the solution,
an iterative coordinate propagation~\cite{propagation}-based embedding is applied.\\
\begin{figure}[htp!]\centering
\includegraphics[scale=0.42]{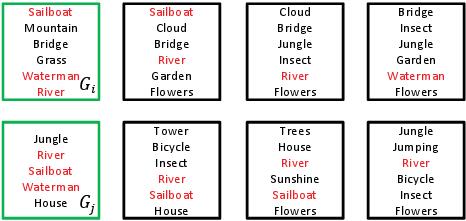}
\caption{An example illustrating the calculation of $l_s$ and
$l_d$ from eight tagged Flickr images. The two green frames denote
images where graphlet $G_i$ and $G_j$ are extracted. In this
example, $c(G_i)\cap c(G_j)=\{\text{Sailboat, Waterman, River}\}$,
$c(G_i)\oplus c(G_j)=\{\text{Mountain, Bridge, Grass, Jungle,
House}\}$, and
=$\{$Sailboat: 5, Mountain: 1, Bridge: 3, Grass: 1, Waterman: 3,
River: 7, Cloud: 2, Garden: 2, Flowers: 5, Jungle: 4, Insect: 4,
House: 3, Tower: 1, Bicycle: 2, Sunshine: 1, Jumping:
1$\}$.}\label{fig3}\end{figure} \indent Based on the manifold
embedding algorithm, we calculate the attribute-level response map
for each image. In particular, given $C$ the number of textual
attributes, we train a multi-class SVM based on the
one-versus-rest scenario. The saliency response of graphlet $G$ to
textual attribute $\mathbf{u}_d$ is calculated based on the
probabilistic output of SVM:
\begin{equation}
p(y(G)\rightarrow
\mathbf{u}_d)=\frac{1}{1+\exp(-\phi_{\mathbf{u}_d}(y(G)))},
\label{eq17a}
\end{equation}
where $y(G)$ is the post-embedding vector corresponding to
graphlet $G$, and $\phi_{\mathbf{u}_d}(\cdot)$ denotes the
classification hyper-plane that separates graphlets belonging to
images with textual attribute $\mathbf{u}_d$
from the remaining graphlets.\\
\indent After obtaining the saliency map to each textual
attribute, we randomly generate 10000 patches in each image. The
patch sizes are tuned as follows: the patch width is tuned from
$[0.1w, 0.9w]$ with a step of $0.01w$, while the patch height is
tuned from $[0.1h, 0.9h]$ with a step of $0.01h$. Note that $w$
and $h$ denote the width and height of an image respectively.
Then, the patch $W$ whose internal graphlets' joint probability
can be maximized is selected,~\textit{i.e.}, \begin{equation}
\max\nolimits_W\prod\nolimits_{G\in W}p(y(G)\rightarrow
\mathbf{u}_d). \label{eq18}\end{equation} Patch $W$ indicates the
visual attribute that is most responsive to textual attribute
$\mathbf{u}_d$. It captures the aesthetically relevant visual
feature and we thus call it aesthlet.
\begin{figure*}[htp!]\centering
\includegraphics[scale=0.72]{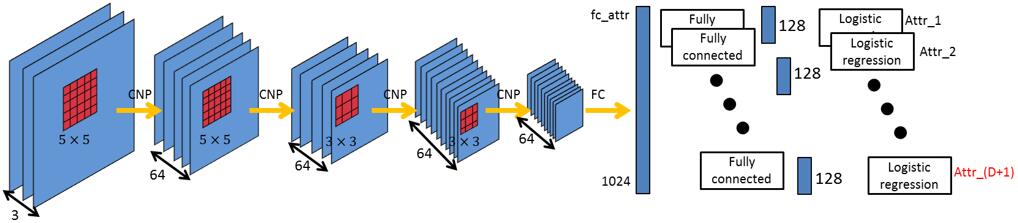}
\caption{A graphical illustration of the proposed deep
architecture (CNP means convolution, normalization and pooling,
and FC denotes the fully-connected layer. The $(D+1)$-th visual
attribute corresponds to the aesthlet describing an entire Flickr
image.)}\label{fig4}\end{figure*}
\subsection{Aesthlet-normalized CNN for Aesthetic Modeling}
We integrate aesthlets into a deep architecture which learns
patch-normalized representations to model visual attributes. After
localizing different aesthlets in a weakly supervised way, we
normalize and feed them into a CNN to extract standard
representations for visual attributes modeling. To this end, we
leverage both the power of CNN to learn discriminative visual
features and the advantage of aesthlets to enhance the CNN
learning by localizing highly aesthetic regions. Noticeably,
although CNN has been successfully applied in a variety of
multimedia tasks, they cannot generalize well when are trained
using small-scale data. Our proposed aesthlets make the learning
process require fewer training examples due to the increased
training set size. This is because a Flickr image usually contains
multiple aesthlets, each of which is treated as a single training example.\\
\indent Starting from a large collection of aesthlet patches, we
resize each of them into $64\times 64$ pixels. Then, we randomly
jetter each patch and flip it horizontally/vertically with
probability 0.5 to improve the generalization, thereby a CNN is
trained to represent each aesthlet. The architecture of the
proposed CNN is elaborated in Fig.~\ref{fig4}. The network
contains four layers,~\textit{i.e.}, convolution, max pooling,
local response normalization and a fully-connected layer with 1024
hidden units. After this, the network branches out one fully
connected layer containing 128 units to describe each visual
attribute. The last two layers are split to form tailored features
for each visual attribute,~\textit{e.g.}, determining whether a
Flickr image is ``brightly-colored'' or ``well-structured'' may
require different visual features. Comparatively, the bottom
layers are shared in order to: 1) reduce the number of parameters,
and 2) take advantage of the common low-layer CNN structure. The
parameters of our developed CNN are adjusted by cross-validation.
Specifically, we first set the entire parameters: the input patch
size, the number of convolutions, and strides exactly the same as
those in Krizhevsky~\textit{et al.}~\cite{imagenet}'s work. Then,
we tune one parameter while leaving the rest unchanged. We set the
tuning parameter by maximizing the accuracy of aesthetic
prediction on
our compiled data set~\cite{zhang_graphlet}.\\
\indent The entire CNN is trained based on the standard
back-propagation~\cite{da3} of the error, combined with a
stochastic gradient decent as a loss function,~\textit{i.e.}, the
sum of the log-loss of each aesthlet from a training image. The
architecture of each CNN layer is shown in Fig.~\ref{fig4} and the
implementation details are elaborated in~\cite{da3}. To
effectively handle noise, we prone to employ aesthlet patches with
high detection accuracies (based on (\ref{eq18})) in the training
stage. In our paper, we did not use the pre-trained network.
Actually, in our experiment, similar to~\cite{imagenet}, we have
tried to adopt a pre-trained network (trained from three aesthetic
images data set: CHUK~\cite{ke}, PNE~\cite{datta}, and
AVA~\cite{ava} respectively) to increase the aesthetic prediction
accuracy.\\\begin{figure}[htp!]\centering
\includegraphics[scale=0.46]{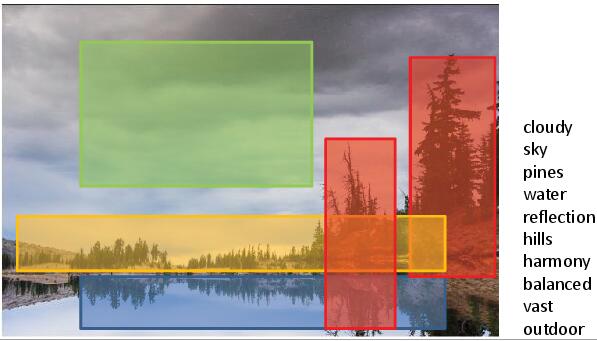}
\caption{Examples comparing the roles of global composition and
aesthlets in describing image aesthetics. The color-tagged regions
indicate aesthlets calculated based on the Flickr tags on the
right.}\label{fig5}\end{figure} \indent As shown in
Fig.~\ref{fig5}, each aesthlet describes the aesthetics of a
Flickr image from a single view. Therefore, we first utilize the
developed CNN with respect to each aesthlet to generate
representative aesthetic feature from a single view. Then, we
combine the representations from multiple views to obtain the
final aesthetic feature describing the entire image. More
specifically, we extract the activations from the $fc\_attr$ layer
in Fig.~\ref{fig4}, which is 1024-dimensional, to describe each
aesthlet. Finally, we concatenate the activations of all the
aesthlets into a feature vector. It is worth emphasizing that, if
an aesthlet does not activate for the image, we simply set the
corresponding feature representation to zero.\\
\indent As exemplified in Fig.~\ref{fig5}, the aforementioned
aesthlet-based CNN exploits regional visual features perceived by
human beings. But the global features also plays an important role
in describing image aesthetics. However, typically aesthlet
patches cannot cover the entire image region. Even worse, in some
degenerated cases, a Flickr image may have very few aesthlets
detected,~\textit{e.g.}, abstract paintings without specific
objects. To deal with this, we also incorporate a CNN whose inputs
are patches covering the whole Flickr image, in order to capture
the global aesthetics of Flickr images. Obviously, patches
covering the whole image can be considered as special aesthlets.
Therefore, the CNN is implemented with the same structure as that
shown in Fig.~\ref{fig4}.\\
\indent Lastly, we concatenate the feature vectors corresponding
to all the aesthlets from each Flickr image into a
$128*(D+1)$-dimensional feature vector, which reflects the
aesthetic feature of the image both locally and globally.
\subsection{Applications of the Deep Aesthetic Features}
We can employ the learned deep aesthetic features to enhance three
applications in multimedia field: photo retargeting,
aesthetics-based image classification and
retrieval.\\\indent \textbf{Image
retargeting:} For image retargeting, a 5-component GMM is adopted
to learn the distribution of deep features calculated from all
training well-aesthetic Flickr images:
\begin{equation}
p(z|\theta)=\sum\nolimits_{l=1}^5
\alpha_l\mathcal{N}(z|\pi_l,\Sigma_l),\label{eq19}
\end{equation}
where $z$ denotes the deeply-learned features, and
$\theta=\{\alpha_l,\pi_l,\Sigma_l\}$ are the GMM parameters.\\
\indent Based on the GMM, we shrink a test image to make its deep
feature most similar to those from the training images. Particularly, we decompose an image into
equal-sized grids. Then, the horizontal (w.r.t. vertical) weight
of grid $\phi$ is calculated as:
\begin{equation}
w_h(\phi)=\max p(z(\phi)|\theta),\label{eq20}
\end{equation}
where $z(\phi)$ denotes the deep aesthetic feature calculated
based on the shrunk image. After obtaining the horizontal (w.r.t.
vertical) weight of each grid, a normalization step is carried out
to make them sum to one:
\begin{equation}
\bar{w}_h(\phi_i)=\frac{w_h{\phi_i}}{\sum\nolimits_i
w_h(\phi_i)},\label{eq20}
\end{equation}
\indent Given the size of the retargeted image $W\times H$, the
horizontal dimension of the $i$-th grid is shrunk to $[W\cdot
\bar{w}_h(\phi_i)]$, where $[\cdot]$ rounds a real number to the
nearest integer. The shrinking operation along the vertical
direction is similar to
that along the horizontal direction.\\
\indent \textbf{Aesthetics-based image classification/retrieval:}
The deep aesthetic feature reflects human aesthetic perception.
They can be used to: 1) identify whether a Flickr image is highly
or low aesthetic, and 2) aesthetics-based image retrieval. The key
of these two applications is a probabilistic model.\\
\begin{figure}[htp!]\centering
\includegraphics[scale=0.44]{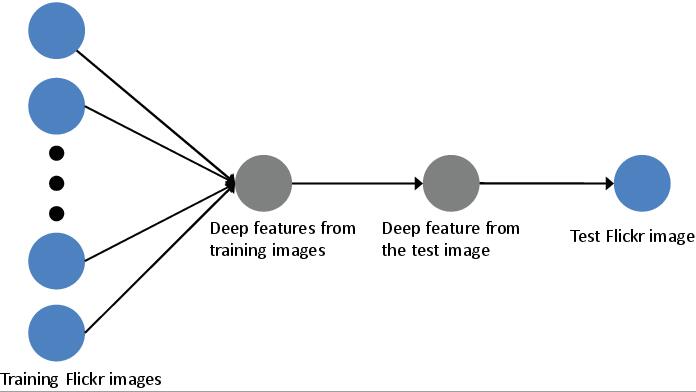}
\caption{A graphical illustration of the probabilistic model for
image aesthetics quantification}\label{fig7}\end{figure} \indent
As shown in Fig.~\ref{fig7}, given a set of training images and a
test one, they are highly correlated through their deep aesthetic
features $z$ and $z^*$ respectively. The probabilistic model
contains four layers. The first layer corresponds to all the
training images $I^1,I^1,\cdots,I^L$ which are aesthetically
pleasing; the second layer denotes all the deep aesthetic features
$z$ learned from the training images; the third layer represents
the deep aesthetic feature $z_*$ extracted from the test image;
and the last layer denotes the test image $I_*$.\\
\indent Naturally, image aesthetics can be quantified as the
amounts of deep aesthetic features that can be transferred from
the training images into the test one. Thus, the aesthetics of a
test image can be quantified as: \begin{eqnarray}
\gamma&&\hspace{-15pt}=p(I_*|I^1,I^1,\cdots,I^L)\nonumber\\
&&\hspace{-15pt}=p(I_*|z_*)\cdot p(z_*|z)\cdot p(z|I^1,I^2,\cdots,I^L),\label{eq25}
\end{eqnarray}
where the probability $p(z_*|z)$ is calculated as:
$p(z_*|z)=\prod\nolimits_{j=1}^L p(z_*|z^j)$. $z_*$ is the deep
aesthetic feature calculated from the test image.
$z^j$ denote the deep aesthetic feature calculated from the $j$-th training image.\\
\indent Following many
algorithms~\cite{prob_fusion,zhang_graphlet}, we define the
similarity between deep aesthetic features as a Gaussian kernel:
\begin{equation} p(z_*|z)\propto
\exp\left(-\frac{||z_*-z||^2}{2\sigma^2}\right).\label{eq27}
\end{equation}
After obtaining the aesthetic score $\gamma$ of a test Flickr
image. If $\gamma>0.5$, then this image is deemed as
``aesthetically pleasing'', and vice versa. Besides, for
aesthetics-based image retrieval, we output images in the database
whose aesthetic scores are similar to that of the query image.

\section{Experimental Results and Analysis}
This section evaluates the performance of the proposed deep
aesthetic feature based on four experiments. The first experiment
visualizes and analyzes the effectiveness of the proposed
aesthlet. Then, we compare the three applications based on our
deeply-learned aesthetic feature with the state-of-the-art. A
step-by-step evaluation of the proposed method is presented
subsequently. The last experiment evaluates the influence of
different parameter settings.\\
\indent All the experiments were carried out on a personal
computer equipped with an Intel i5-2520M CPU and 8GB RAM. The
algorithm was implemented on the Matlab 2012 platform.
\subsection{Descriptiveness of the Proposed Aesthlet}
\begin{figure*}[htp!]\centering
\includegraphics[scale=0.68]{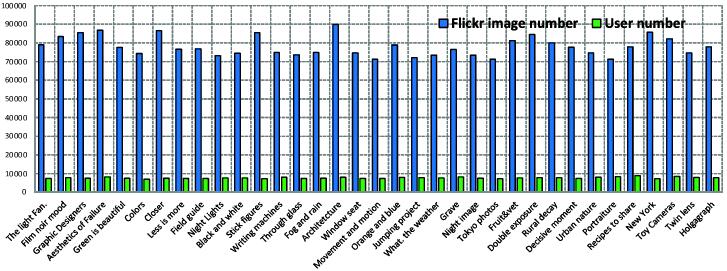}
\caption{The 35 Flickr groups (the horizontal axis) and the number
of users (the vertical axis) in each of group}\label{fig8}\end{figure*}

As the input to our deep architecture, aesthlets are image patches
which are aesthetically pleasing and correspond to the textual
attributes in each Flickr image. In this experiment, we first
visualize the extracted aesthlets on a subset of our own complied
Flickr image~\cite{icme} dataset. Then, a comprehensive user study
based on 132 participants is conducted to
evaluate the descriptiveness of our proposed aesthlets.\\
\indent \textbf{Dataset Compilation:} We spent significant time,
effort and resources to crawl photos from 35 well-known groups
form Flickr. For each group, we collected $70,000\sim 90,000$
photos from nearly 7,400 Flickr users. The statistics of our
dataset is shown in Fig.~\ref{fig8}. For different Flickr groups,
the numbers of photos belonging to each user varies from 10 to
220. We rank the Flickr images from each group based on the
aesthetic measure by Zhang~\textit{et al.}~\cite{zhang_graphlet}.
The top 10\% highly aesthetic photos
constitute the image set for our experiment.\\
\begin{figure}[htp!]\centering
\includegraphics[scale=0.58]{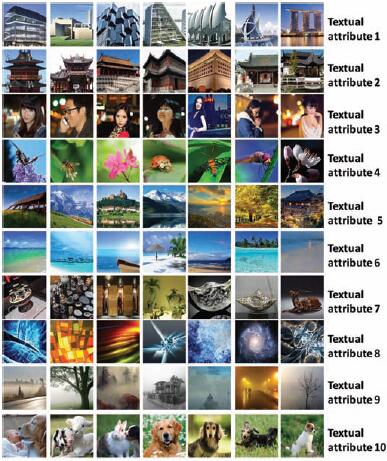}
\caption{Visualized aesthlets corresponding to each textual
attribute} \label{fig9}\end{figure} \indent We set the number of
textual attributes $D=10$ and calculate the corresponding
aesthlets according to (\ref{eq18}). The representative aesthlets
describing each textural attribute is presented in
Fig.~\ref{fig9}. The following observations can be made:
\begin{itemize}
\item The learned textural attributes are representative to each
Flickr image, as the corresponding visual attributes can
accurately detect aesthetically pleasing regions in each image.
Moreover, we notice that each Flickr image is associated with
fewer than three textural attributes. This is achieved by the
sparse LSA and can accelerate the aesthetic modeling remarkably.
\item Our graphlet-based weakly supervised learning algorithm
effectively maps the textual attributes to the corresponding
aesthelts. The is because graphelts can seamlessly capture objects
with various shapes, and the weakly supervised algorithm can be
solved analytically and efficiently. \item We notice that
aesthlets corresponding to the same textual attribute have highly
similar semantics, such as ``modern architecture''.
Simultaneously, aesthlets corresponding to different textual
attributes have distinguishable semantics, such as ``scenery'' and
``beach''. This demonstrates the effectiveness of our aesthlets
discovering mechanism, since the ratio between inter-attribute
scatter and intra-attribute scatter is prone to be maximized.
\end{itemize} As aesthlets reflect human aesthetic perception, it
is infeasible to measure their descriptiveness quantitatively. In
this experiment, we conduct a user study to evaluate them
qualitatively. This strategy was also adopted
in~\cite{zhang_graphlet} for comparing the aesthetic quality of
multiple cropped photos. We invited 132 participants and most of
them are master/Phd students from computer sciences department.
For each aesthlet, we asked each participant to indicate its
preference and discrimination scores. The former reflects the
attractiveness of this aesthlet, while the latter shows whether
this aesthelt captures a unique type of aesthetics. We set of
number of textual attributes $D$ to 10, 20, 40, 80 and 160
respectively. Then, the average preference and discrimination
scores of aesthlets are calculated and reported in
Table~\ref{tab1}. As can be seen, the preference scores under
different values of $D$ are above 0.9, which shows that the
extracted aesthlets can accurately localizing visually attractive
regions. Comparatively, the discrimination of aesthlet decreases
dramatically when the value of $D$ increases. The reason might be
that the number of latent aesthetic types is much smaller than the
number of textual attributes $D$.\begin{table}\centering
\caption{Average preference and discrimination scores of aesthlets
by varying $D$}\begin{tabular}{c|c|c|c|c|c}\hline
 &$D=10$ &$D=20$ &$D=40$ &$D=80$ &$D=160$ \\\hline
Preference &0.9434&0.9221&0.9121&0.9098&0.9002\\\hline
Discrimination &0.9432&0.8956&0.8211&0.7676&0.6545 \\\hline
\end{tabular} \label{tab1}\end{table}
\subsection{Comparative Study}
\begin{figure*}[htp!]\centering
\includegraphics[scale=0.67]{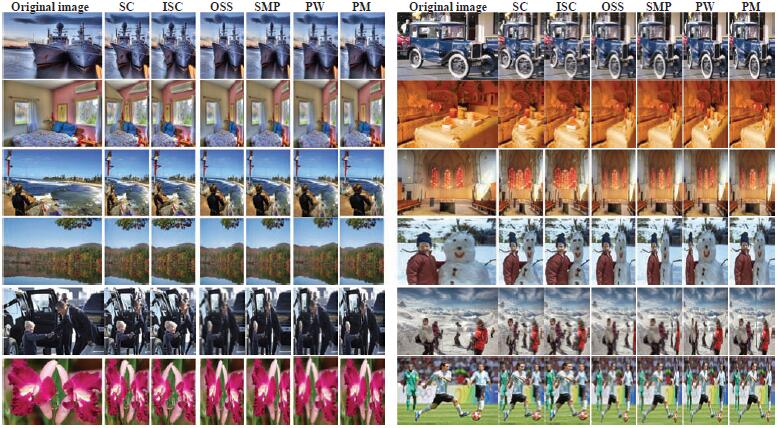}
\caption{Photos retargeted based on different algorithms}
\label{fig10}\end{figure*}

\begin{figure}[htp!]\centering
\includegraphics[scale=0.56]{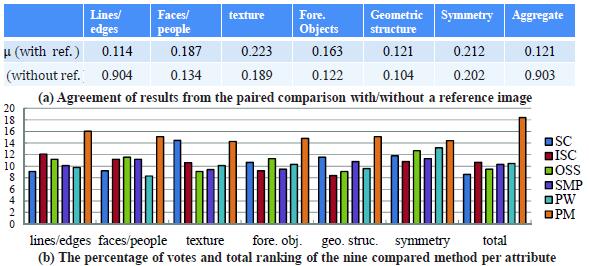}
\caption{A detailed analysis of the comparative retargeted photos
in Fig.~\ref{fig10}} \label{fig11}\end{figure}

This subsection evaluates the three applications based on our
deeply-learned aesthetic features: image retargeting, aesthetics-based image classification and retrieval.\\
\indent \textbf{Image retargeting:} Fig.~\ref{fig10} compares the proposed
method (PM) against several representative state-of-the-art
approaches, including: seam carving (SC)~\cite{sc} and
its improved version (ISC)~\cite{isc}, optimized scale-and-sketch
(OSS)~\cite{optimal} and saliency-based mesh parametrization
(SMP)~\cite{mesh_para}, and the patch-based wrapping (PW)~\cite{patch}.
Resolution of the resulting images is fixed to $640\times 960$. The experimental images
are all from the RetargetMe dataset~\cite{retargetme} \\
 \indent In order to make the evaluation
comprehensive, we adopt a paired comparison-based user study to
evaluate the effectiveness of our proposed retargeting method. In
the paired comparison, each participant is presented with a pair
of retargeted photos from two different approaches, and required
to indicate a preference as of which one they would choose for an
iPhone wallpaper. In our user study, the participants are 40
amateur/professional photographers. As shown in Fig.~\ref{fig10},
compared with its competitors, our approach well preserves the
semantically important objects in the original photos, such as the
warship and the bubble car. In contrast, the compared retargeting
methods sometimes shrink the semantically important objects, such
as the viewer, and boy and the football player. Even worse, SC and
its variant ISC, as well as OSS may result in visual
distortions.\\
\indent In addition, we present an in-depth analysis of the user
study on the 12 sets of retargeted photos in Fig.~\ref{fig10},
which is inspired by the comparative study of retargeting
algorithms in~\cite{retargetme}. We invited 132 volunteers and
most of them are master/Phd students from computer sciences
department. For each volunteer, we asked them two questions. 1)
Whether each visualized aesthlet is preferred by him/her and
further indicate its aesthetic level, which is a real number
ranging from 0 to 1. 0 indicates the lowest aesthetics while 1
denotes the highest aesthetics. 2) Whether each visualized
aesthlet corresponding to a textual attribute is distinguishable
from those corresponding to the rest textural attribute, and
further indicate its discriminative level. 0 indicates the lowest
discrimination while 1 denotes the highest discrimination. After
collecting the preference and discrimination scores of all the
aesthlets, we average them. Obviously, these two measures can show
the effectiveness of our proposed method. First, we evaluate the
degree of agreement when the volunteers vote for their favorite
retargeted images, where a high disagreement reflects the
difficulty in decision making. In our experiment, we use the
coefficient of agreement defined by Kendall and
Babington-Smith~\cite{r16}. The coefficients over all the 12
retargeted images are shown in the last column of
Fig.~\ref{fig11}(a). Besides, we also collect the volunteers'
votes on each attribute of the four sets of retargeted photos. As
shown from the second column to the seventh column in
Fig.~\ref{fig11}(a), the volunteers are highly agreable on the
face/people, the texture, and the symmetry because these
attributes are well preserved by our retargeting model. Then, we
present the votes on each attribute based on the 12 retargeted
images. As shown in Fig.~\ref{fig11}(b), the proposed method
receives the most votes on all the attributes consistently.\\
 \indent
\textbf{Aesthetics-based image classification:} We compare our
approach with five image aesthetics evaluation methods. The
compared methods include three global feature-based approaches
proposed by Dhar~\textit{et al.}~\cite{dhar}, Luo~\textit{et
al.}~\cite{luo2} and Marchesotti~\textit{et
al.}~\cite{marchesotti} respectively, two local patch
integration-based methods proposed by Cheng~\textit{et
al.}~\cite{cheng} and Nishiyama~\textit{et al.}~\cite{nishiyama2}
respectively, and the CNN-based aesthetic model by Lu~\textit{et
al.}. In~\cite{da9}, Lu~\textit{et al.} proposed a novel framework
to predict image style, aesthetics, and quality. The key technique
is a deep network which learns the fine-grained details from
multiple image patches, where multi-patch aggregation functions
can be learned as part of neural network training. Our proposed
aesthlet-based image aesthetic model is significantly different
from Lu~\textit{et al.}'s method, which is conducted in an
unsupervised way. In our approach, the image patches
(\textit{i.e.}, aesthlets) are detected using a weakly-supervised
learning algorithm, wherein the weak labels are latent topics
detected from the $M$-dimensional augmented frequency vector.\\
\begin{table} \centering
\caption{Comparative aesthetic quality prediction accuracies}
\begin{tabular}{l|c|c|c}\hline
&CUHK   & PNE &AVA\\\hline\hline
Dhar \textit{et al.} &0.7386&0.6754 &0.6435\\\hline
Luo \textit{et al.} & 0.8004 & 0.7213&0.6879 \\\hline
Marchesotti \textit{et al.} (FV-Color-SP)&0.8767  &0.8114 &0.7891\\\hline
Cheng \textit{et al.} &   0.8432& 0.7754 &0.8121\\\hline
Nishiyama \textit{et al.}  &  0.7745 &0.7341 &0.7659\\\hline
Lu \textit{et al.}  &  0.8043 &0.8224 &0.8213\\\hline
The proposed method   & \textbf{0.8879} &\textbf{0.8622} &\textbf{0.8465}\\\hline\end{tabular}\label{tab2}
\end{table}\indent In the comparative study, we
observe that the source codes of the five baseline methods are
unavailable and some experimental details are not provided. This
makes it difficult to implement them exactly. In our
implementation, we try to strengthen some components of the
baseline methods. The following settings are employed. For Dhar's
approach, we use the public codes from Li~\textit{et
al.}~\cite{feifei} to extract the attributes from each photo.
These attributes are combined with the low-level features proposed
by Yeh~\textit{et al.}~\cite{yeh} to train the aesthetic
classifier. For Luo~\textit{et al.}'s approach, not only are the
low-level and high-level features in their publication
implemented, but also the six global features from
Getlter~\textit{et al.}~\cite{on_comb}'s paper are utilized to
strengthen the aesthetic prediction ability. For
Marchesotti~\textit{et al.}'s approach, similar to the
implementation of Luo~\textit{et al.}'s method, the six additional
features are also adopted. For Cheng~\textit{et al.}'s algorithm,
we implement it as a reduced version of our method, where only
2-sized graphlets are utilized for aesthetics measure. It is worth
emphasizing that, for the probabilistic aesthetics evaluation
models proposed by Cheng~\textit{et al.}, Nishiyama~\textit{et
al.} method, and us, if the aesthetic score is higher than 0.5,
then this image is categorized as highly aesthetic, and vice versa.\\
\indent We present the aesthetics prediction accuracies on the
CUHK, PNE and AVA in Table~\ref{tab2}. The image tags from the
three datasets are assigned by ourselves. We hired 50 master/Phd
students from our department to conduct the label annotation task.
Each student spent $2\sim 4$ hours to annotate $500\sim 700$
images. All the images from the three data sets: the CHUK (12,000
images), the PNE (nearly 1700 images), and the AVA (about 25,000
images) are annotated. As shown, our approach outperforms
Marchesotti~\textit{et al.}'s approach by nearly 2\%, and exceeds
the rest of the compared methods by over 6\%,
The results demonstrate the clear advantage of our method.\\
\indent \textbf{Aesthetics-guided image retrieval:} We adopt the
precision rate~\cite{pr} to evaluate the performance of image
retrieval based on our deeply-learned aesthetic features.
Precision denotes the ratio of the number of relevant images (to
the user) to the scope $S$, which is specified by the number of
top-ranked images. In the experiment, we observe that it is
sufficient to display 30 retrieved images on a screen. Presenting
more images may decrease the quality of the presented images.
Therefore, we set $S=30$ throughout the experiment. The
experimental dataset is our own crawled large-scale Flickr
images from 35 groups.\\
\indent In the current image retrieval systems, typically the
query image is not exist in the image database. To handle this
problem, we randomly select 30 images from each Flickr group as
query images, while the rest are treated as the database for
retrieval. Notably, the precision rate is calculated by
averaging the results over the $30*35=1050$ queries.\\
\indent To alleviate computational burden, we select the most
informative images from the top 400 images. Relevance feedback is
adopted to bridge the semantic gap in the retrieval system. The
relevance feedback system contains a two-stage process: i) users
are shown 30 images deemed most informative by $SVM_{active}$; ii)
users give relevance feedback; iii) (i) is repeated taking into
account feedback from ii). Users are asked to make relevance
judgements toward the queries results. Thereafter, the feedback
information is utilized to re-rank the images in the database.
$SVM_{active}$ is used as the relevance feedback algorithm. It
provides users with the most informative
images with respect to the ranking function.\\
\indent We compare our deep features with well-known aesthetic
descriptors proposed by Marchesotti~\textit{et
al.}~\cite{marchesotti}, Cheng~\textit{et al.}~\cite{cheng},
Lu~\textit{et al.}~\cite{da9} and Champbell~\textit{et
al.}~\cite{da10} respectively. As shown in Table~\ref{tab3}, in
most of the 35 Flickr groups, our deeply-learned aesthetic feature
achieves the best precision. We also observe that, although
Marchesotti~\textit{et al.}'s aesthetic features is simple, its
performance is competitive. \begin{table}\footnotesize\centering
\caption{Precision at top 30 returns of the five compared
aesthetic features. The highest precision is in bold for each
Flickr group.}\begin{tabular}{c|c|c|c|c|c}\hline Flickr group
&March. &Cheng &Lu &Champ. &Ours \\\hline\hline The light Fan.
&0.34 &0.33 &0.43 &0.31 &\textbf{0.53}\\\hline Film noir Mood
&0.16 &0.19 &0.23 &0.16 &\textbf{0.26}\\\hline Graphic designers
&\textbf{0.33} &0.27 &0.31 &0.24 &0.29\\\hline Aesthetics failure
&0.11 &0.09 &0.15 &0.12 &\textbf{0.19} \\\hline Green is beautiful
&0.14 &0.13 &0.16 &0.12 &\textbf{0.22}\\\hline Colors
&\textbf{0.34} &0.25 &0.28  &0.21 &0.31 \\\hline Closer &0.43
&0.41 &0.45 &0.36 &\textbf{0.51}\\\hline Less is more &0.26 &0.22
&0.25 &0.21 &\textbf{0.34} \\\hline Field guide &0.34 &0.31 &0.33
&0.28 &\textbf{0.41}\\\hline Night lights &0.16 &0.14 &0.16 &0.13
&\textbf{0.19}\\\hline Black and white &\textbf{0.21} &0.18 &0.19
&0.15 &0.18\\\hline Stick figure &0.43 &0.41 &0.34
&0.37&\textbf{0.47} \\\hline Writing mach. &0.67 &0.64 &0.68 &0.56
&\textbf{0.76}\\\hline Through glass &0.09 &0.09 &0.07 &0.05
&\textbf{0.12} \\\hline Fog and rain &0.21 &0.20 &0.18 &0.19
&\textbf{0.25} \\\hline Architecture &0.54 &0.53 &0.54 &0.46
&\textbf{0.59} \\\hline Window seat &0.42 &0.39 &0.40 &0.36
&\textbf{0.46}\\\hline Movement &0.35 &0.31 &0.33 &0.31
&\textbf{0.37} \\\hline Orange and blue &0.18 &0.16 &0.18 &0.13
&\textbf{0.21} \\\hline Jump Project &0.26 &0.24 &0.19& 0.22
&\textbf{0.31}\\\hline What. the weather &0.14 &0.12 &0.11 & 0.10
&\textbf{0.17}\\\hline Grave &0.28 &0.24 &0.26 &0.21
&\textbf{0.31} \\\hline Night images &0.14 &0.16 &0.12 &0.11
&\textbf{0.19}\\\hline Tokyo photos &0.24 &0.21 &0.19 &0.19
&\textbf{0.26}\\\hline Fruit\&vet &0.65 &0.62 &0.61 &0.62
&\textbf{0.69}\\\hline Double exposure &0.28 &0.23 &0.24 &0.19
&\textbf{0.32}\\\hline Rural decay &0.25 &0.23 &0.24 &0.19
&\textbf{0.28}\\\hline Decisive moment &0.24 &0.22 &0.19 &0.17
&\textbf{0.27}\\\hline Urban nature &\textbf{0.32} &0.26 &0.27
&0.23 &0.29\\\hline Portraiture &0.54 &0.46 &0.48 &0.47
&\textbf{0.59}\\\hline Recipes to share &0.26 &0.22 &0.24 &0.21
&\textbf{0.31}\\\hline New York &0.27 &0.23 &0.25 &0.21
&\textbf{0.31}\\\hline Toy cameras &0.37 &0.31 &0.31 &0.30
&\textbf{0.42}\\\hline Twin lens &0.26 &0.22 &0.24 &0.21
&\textbf{0.29}\\\hline Holgagraph &0.33 &0.28 &0.25 &0.21
&\textbf{0.38}\\\hline\hline Average &0.28 &0.23 &0.21 &0.19
&\textbf{0.33}\\\hline
\end{tabular} \label{tab3}
\end{table}
\subsection{Step-by-step Model Justification}
This experiment validates the effectiveness of the three key
components in our deep aesthetic feature learning framework: 1)
sparsity-constrained textual attributes discovery; 2) weakly
supervised visual attributes localization; and 3) the
aesthlet-normalized CNN training. To empirically demonstrate the
effectiveness and inseparability of these components, we replace
each component by a functionally reduced counterpart and report
the corresponding performance. We focus on the application of
aesthetics-based image classification. \\
\indent \textbf{Step 1:} To demonstrate the usefulness of the
sparse textual discovery, three experimental settings are used to
weaken the adopted sparse LSA. First, we abandon the sparsity
constraint in (4). Afterward, we replace the sparse LSA by the
well-known linear discriminate analysis (LDA) and principle
component analysis (PCA) respectively. We present the calculated
image classification accuracies in Table~\ref{tab4}. As shown,
removing the sparsity constraint results in an accuracy decrement
of 5.2\% on average. Moreover, either LDA nor PCA cannot optimally
discover the latent textual attributes, since their performances
lag behind ours by over 10\%.\\
\indent \textbf{Step 2:} To evaluate the effectiveness of the
weakly supervised visual attributes learning, three experimental
settings are adopted. We first replace the graphlet-based object
detection by the objectness measure~\cite{objectness} and the
part-based object detector~\cite{part_based} respectively. Then,
we replace the superpixel-based spatial pyramid by the standard
grid-based one. As shown in Table~\ref{tab4}, on average, both the
objectness and part-based detector cause a nearly 1.2\% accuracy
decrement. Noticeably, the training process of them requires
manually annotated windows, which might be computationally
intractable. Besides, grid-based spatial pyramid severely decrease
the performance of aesthetics-based image classification. The
reason is, compared with the superpixel-based graphlets,
grid-based graphlets cannot fit the various
object shapes optimally.\\
\indent \textbf{Step 3:} We study the performance of our developed
CNN. We remove the fully-connected layer and observe that the
aesthetics-based classification reduces by 4\% on average. This
result reflects the necessity to explore the common low-layer CNN
structure. Additionally, we abandon the global image visual
attribute. On average, a decrement of nearly 7\% is observed. This
shows the importance of explicitly modeling global image
configurations in aesthetic modeling. \begin{table} \centering
\caption{Performance decrement by varying the experimental
settings in each component}
\begin{tabular}{l|c|c|c}\hline
&CUHK   & PNE &AVA\\\hline\hline Step 1:~Sparsity &5.4\% & 4.9\% &
5.2\%\\\hline Step 1:~Remove LSA &7.5\% & 6.7\% & 7.1\%\\\hline
Step 1:~LDA &9.6\% &10.9\%&10.3\% \\\hline Step 1:~PCA &11.3\%
&9.5\% &10.7\%\\\hline Step 2:~Objectness & 1.3\% &1.0\%
&1.1\%\\\hline Step 2:~Part-based & 1.1\% &1.1\% &1.4\%\\\hline
Step 2:~Superpixel & 7.9\% & 8.4\% &7.3\%\\\hline Step
3:~Fully-connected & 4.3\%& 3.8\% & 4.0\%\\\hline Step 3:~Remove
global descriptor &8.4\%& 6.2\%&6.7\%\\\hline Step 3:~Only global
descriptor &12.3\%& 14.1\%&12.7\%\\\hline
\end{tabular}\label{tab4}\end{table}
\subsection{Parameter Analysis}
In this experiment, we evaluate the influence of important
parameters in our proposed aesthetic model: 1) the number of
textual attributes $D$, 2) the regularizer weight $\lambda$ in
sparse LSA and 3) the structure of our designed
CNN.\\\begin{figure}[htp!]\centering
\includegraphics[scale=0.68]{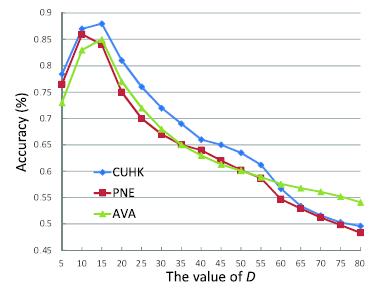}
\caption{Aesthetics-based image classification accuracies by
varying $D$ on the CUHK, PNE and AVA datasets}
\label{fig12}\end{figure} \indent First, we tune the value of $D$
from five to 80 with a step of five and report the corresponding
performance of aesthetics-based image classification. As shown in
Fig.~\ref{fig12}, on all the three datasets, the best accuracies
are achieved when $D$ is set to 10 or 15. This indicates there are
about $10\sim 15$ latent semantic categories from each of the
three datasets. Second, we choose the value of $\lambda$ from
$\{0.5, 0.2, 0.1,0.05,0.001\}$ and report the corresponding
accuracies. As shown in Table~\ref{tab5}, we notice that the best
accuracies on all the three datasets are achieved when
$\lambda=0.1$. Next, we testify the effectiveness of our developed
five-layered CNN. We change our CNN to a four- and six-layered CNN
respectively. It is noticeable that the aesthetics-based image
classification accuracies decreased by 4.3\% and 6.7\%
respectively. Actually, in our implementation, the five CNN layers
are validated by cross validations. Lastly, we preserve only the
global descriptor for image aesthetics prediction, where the
CNN-based descriptor is abandoned. As shown in the last row of
Table~\ref{tab5}, aesthetics-based image classification using only
global descriptor results in an accuracy decrement of 13\% on
average. This observation clearly demonstrates the necessity of
exploiting local descriptors in aesthetic modeling.
\begin{table} \centering \caption{Aesthetics-based
image classification accuracies under different values of
$\lambda$}
\begin{tabular}{l|c|c|c|c|c}\hline
Dataset&$\lambda=0.5$   & $\lambda=0.2$   &$\lambda=0.1$
&$\lambda=0.05$ &$\lambda=0.001$  \\\hline\hline CHUK
&0.7612&0.8133&0.8879&0.6454&0.7687 \\\hline PNE
&0.7453&0.8231&0.8622&0.7121&0.7453\\\hline AVA
&0.8113&0.8214&0.8465&0.7376&0.7786\\\hline
\end{tabular}\label{tab5}\end{table}

\section{Conclusions and Future Work}
Perceptually aesthetic model is an important topic in multimedia
field~\cite{tmm3,tmm4,tmm5}. This paper proposes a CNN framework
to hierarchically model how humans perceive aesthetically pleasing
regions in a Flickr image. We first calculate a compact set of
textual attributes from tagged Flickr images using a
sparsity-constrained LSA. Then, a weakly supervised learning
paradigm projects the textual attributes onto the corresponding
aesthlets in each image. These aesthlets capture visually
attractive image regions and are deployed to train a CNN to mimick
human aesthetic perception. Based on the CNN, we represent each
Flickr image by a set of deeply-learned aesthetic features, which
can enhance a series of multimedia applications,~\textit{e.g.},
image retargeting, aesthetics-based
image classification and retrieval.\\
\indent In the future, this work will be extended to a more
comprehensive deep architecture that encodes auxiliary visual cues
such as exposure, contrast and symmetry, which might also be
contributive to image aesthetic modeling.

\bibliographystyle{model1a-num-names}

\end{document}